\documentclass[runningheads]{llncs}

\usepackage{eccv}

\usepackage{eccvabbrv}

\usepackage{graphicx}
\usepackage{booktabs}

\usepackage[accsupp]{axessibility}  

\usepackage{epsfig}
\usepackage{amsmath}
\usepackage{amssymb}
\usepackage{multirow}
\usepackage{colortbl}
\definecolor{mygray}{gray}{.9}
\usepackage{marvosym}

\usepackage{hyperref}

\usepackage{orcidlink}

\begin{document}

\title{IRGen: Generative Modeling for Image Retrieval} 


\author{Yidan Zhang\inst{2} \and
Ting Zhang\inst{1}\textsuperscript{*} \and
Dong Chen\inst{3} \and
Yujing Wang\inst{3} \and
Qi Chen\inst{3} \and \\
Xing Xie\inst{3} \and
Hao Sun\inst{3} \and
Weiwei Deng\inst{3} \and
Qi Zhang\inst{3} \and
Fan Yang\inst{3} \and
Mao Yang\inst{3} \and \\
Qingmin Liao\inst{5} \and
Jingdong Wang\inst{4}\and
Baining Guo\inst{3}}

\authorrunning{Y.~Zhang et al.}


\institute{
\textsuperscript{1}Beijing Normal University \hspace{0.1in}
\textsuperscript{2}The University of Tokyo \hspace{0.1in}
\textsuperscript{3}Microsoft  \hspace{0.1in}
\textsuperscript{4}Baidu \\
\textsuperscript{5}Shenzhen International Graduate School, Tsinghua University
}

\maketitle

{\let\thefootnote\relax\footnote{{*Corresponding author: tingzhang@bnu.edu.cn.}}}

\begin{abstract}
  While generative modeling has become prevalent across numerous research fields, its integration into the realm of image retrieval remains largely unexplored and underjustified. In this paper, we present a novel methodology, reframing image retrieval as a variant of generative modeling and employing a sequence-to-sequence model. This approach is harmoniously aligned with the current trend towards unification in research, presenting a cohesive framework that allows for end-to-end differentiable searching. This, in turn, facilitates superior performance via direct optimization techniques. The development of our model, dubbed IRGen, addresses the critical technical challenge of converting an image into a concise sequence of semantic units, which is pivotal for enabling efficient and effective search. Extensive experiments demonstrate that our model achieves state-of-the-art performance  on three widely-used image retrieval benchmarks as well as two million-scale datasets, yielding significant improvement compared to prior competitive retrieval methods. In addition, the notable surge in precision scores facilitated by generative modeling presents the potential to bypass the reranking phase, which is traditionally indispensable in practical retrieval workflows. The code is publicly available at \href{https://github.com/yakt00/IRGen}{https://github.com/yakt00/IRGen}.

\end{abstract}

\section{Introduction}
\label{sec:intro}

Generative modeling has made significant progress in a wide range of tasks including machine translation\cite{vaswani2017attention}, conversational modeling\cite{devlin2018bert, brown2020language, ouyang2022training, adiwardana2020towards}, image captioning\cite{yu2022coca}, image classification\cite{chen2020generative}, text-to-image synthesis\cite{ ramesh2022hierarchical, ding2022cogview2}, and many more. Originating from language and then expanding to other modalities with specially designed tokenizers,
such a universal modeling approach provides a promising direction for unifying different tasks into a versatile pretrained model, which has attracted widespread attention\cite{alayrac2022flamingo, wang2022image, ouyang2022training, li2023blip}. Yet its potential in image retrieval has been unexplored. This paper aims to take the unified trend one step further and investigates generative modeling for image retrieval. 

\nocite{radford2019language,radford2018improving,ramesh2021zero,wu2022nuwa,yu2022coca}

A practical retrieval system generally consists of two stages: 
feature representation learning\cite{el2021training, liu2022nonuniform, lee2022correlation, yang2021dolg} and Approximate Nearest Neighbor (ANN) search\cite{babenko2014inverted, johnson2019billion, guo2020accelerating, chen2021spann}. 
Most image retrieval methods focus only on one individual  stage while ignoring the fact that both stages are inherently and deeply connected in actual service. Thus, the practical system often requires careful per-task hyperparameter tuning to make the most out of the coordination of the feature extraction and ANN search.
While some efforts have been dedicated to joint learning of embedding and compression (or indexing), such as deep quantization / hashing\cite{zhang2021joint,zhan2021jointly,10.5555/3016387.3016390,8953697,chen2020differentiable}, they still rely on a post process conducting non-exhaustive search.

\nocite{cao2020unifying,simeoni2019local,tan2021instance,teichmann2019detect,jayaram2019diskann,ren2020hm}

In this paper, we cast image retrieval as a form of generative modeling and utilize standard Transformer architecture, as in GPT\cite{brown2020language, radford2019language, radford2018improving}, to enable end-to-end differentiable search. Our model, IRGen, 
is a sequence-to-sequence model 
that directly generates identifiers corresponding to the given query image's nearest neighbors. 
Specifically, the model takes a query image as input 
and autoregressively predicts discrete visual tokens, which are considered as the identifier of an image.
The predicted visual tokens are supposed to point to the query image's nearest neighbor. 
As such,
IRGen 
can be trained directly from the final retrieval target starting with raw images.
Two fundamental concerns need to be addressed to enable efficient and effective image retrieval using generative modeling.
First, autoregressive generative modeling is notable for its slow inference speed due to the inherently sequential nature, thus the run-time cost for retrieval increases quadratically with the length of the input sequence.
Second, it is particularly difficult to model the semantic relationship between the identifiers if we drastically shortened the image identifier. Therefore, a semantic tokenizer specially designed for image retrieval is an immediate problem. 

We observe that existing image tokenizers\cite{van2017neural,lee2022autoregressive}, intended for image generation, are not suitable for image retrieval task, leading to poor performance as we will demonstrate in our experiments.
We hence propose 
several key ingredients that first inject semantic information by applying image-level supervision rather than pixel-level reconstructive supervision, then generate dependent tokens in a sequence by leveraging the recursive property of residual quantization, and lastly ensure fast inference speed by tremendously reducing the length of the sequence via exploiting the global feature instead of spatial patch embeddings. 
Afterwards, we intentionally adopt the standard Transformer architecture so that we could scale up the model using existing techniques and infrastructures.

The proposed IRGen achieves state-of-the-art performance across a diverse range of image retrieval datasets, owing to its end-to-end differentiable search capability. A comprehensive empirical study over three image retrieval benchmarks proves that our model is superior to prior state-of-the-art baselines, sometimes even surpassing linear scan search methods. For instance, compared with the best baseline method, including linear scan search, our model achieves remarkable improvement, such as a 20.2\% increase in precision@10 on In-shop Clothes\cite{liu2016deepfashion}, a 6.0\% boost in precision@2 on CUB200\cite{wah2011caltech} and a 2.4\% enhancement in precision@2 on Cars196\cite{krause20133d}.
To assess the scalability of our model, we further experiment on million-level datasets, namely ImageNet\cite{deng2009imagenet} and Places365\cite{zhou2017places}, and consistently demonstrate superior performance in these challenging scenarios.

It is our belief that generative models have the potential to redefine the landscape of image retrieval. The application of generative modeling to image retrieval tasks represents an exciting opportunity to unify information retrieval across various modalities. At a technical level, our model, IRGen, effortlessly bridges the gap between feature representation learning and approximate non-exhaustive search (bypassing vector compression), creating an end-to-end differentiable framework that enables direct optimization based on ultimate retrieval goal. Furthermore, the entire framework is conceptually straightforward, with all components relying on the standard Transformer architecture, renowned for its remarkable scalability\cite{du2022glam, chowdhery2022palm, shoeybi2019megatron, xu2021gspmd}. 

Our contributions are concluded as:
\begin{itemize}
    \item[*] To the best of our knowledge, our work represents the first effort in employing a sequence-to-sequence generative model to address image retrieval in a fully end-to-end manner that directly generates the nearest neighbor’s identifier given the query.
    \item[*] We introduce a semantic image tokenizer explicitly designed for image retrieval to obtain image identifiers, facilitating autoregressive training. It is essential to note that this tokenizer differs significantly from previous quantization models, as our identifier solely represents the image and is not employed for distance computation, as is the case in quantization.
    \item[*] Extensive experiments show that our approach exhibits substantial improvements, achieving new state-of-the-art results. We have demonstrated its scalability to million-scale datasets and provided throughput number, showcasing its near-real-time performance.
\end{itemize}

\section{Related Work}
\noindent \textbf{Image retrieval.}
\nocite{smeulders2000content,lew2006content,liu2007survey,zhang2013image,alzu2015semantic,li2016socializing,zhou2017recent,simonyan2014very,razavian2016visual,sharif2014cnn,noh2017large,yuan2020defense,jun2019combination,noh2017large,xiao2016learning, lagunes2020centroids, wen2016discriminative,bai2019re,cao2020unifying,norouzi2013cartesian,wang2014optimized,babenko2014additive,martinez2014stacked,qiu2017deep,cao2017collective,erin2015deep,bentley1990k,jayaram2019diskann,ren2020hm,andoni2008near,cho2020x,gulrajani2016pixelvae, de2022multilingual,de2021highly,bevilacqua2022autoregressive}
Traditionally, hand-crafted features are heuristically designed to describe the image content based on its color\cite{wengert2011bag,wang2011interactive}, texture\cite{park2002fast,wang2014content} or shape\cite{cao2011edgel}.
Typical features include GIST\cite{siagian2007rapid}, SIFT\cite{lowe1999object}, SURF\cite{bay2006surf}, VLAD\cite{jegou2010aggregating} and so on. 
Recent years have witnessed the explosive research on deep learning based features trained over labeled images. Besides the evolvement of the network architecture designs\cite{krizhevsky2017imagenet,he2016deep,vaswani2017attention}, numerous efforts\cite{wieczorek2020strong,el2021training} have been dedicated to various loss functions including classification loss\cite{zhai2018classification, zhou2019omni}, triplet loss\cite{yuan2020defense}, contrastive loss\cite{jun2019combination,el2021training}, center loss\cite{wieczorek2020strong} and so on. 
The similarity is calculated through distance measure between features or evaluated through re-ranking techniques\cite{revaud2019learning}.
Another different line of research centers on approximate nearest neighbor search 
to speed up the search process, accepting a certain level of compromise in search accuracy. One way is to enable fast distance computation through hashing or quantization techniques such as LSH\cite{indyk1998approximate}, min-Hash\cite{chum2008near}, ITQ\cite{gong2012iterative}, PQ\cite{jegou2010product}, and many others\cite{ge2013optimized, wang2018composite,zhu2016deep}.
The other way is to reduce the number of distance comparison by retrieving a small number of candidates. Typical methods include
partition-based indexing\cite{babenko2014inverted,xia2013joint} that partitions the feature space into non-overlapping clusters and graph-based indexing\cite{jayaram2019diskann} that builds a neighborhood graph with edges connecting similar images. To improve the recall rate while ensuring fast search speed, hierarchical course-to-fine strategy\cite{malkov2018efficient} has been the popular choice that the retrieved candidates are refined level by level. Additionally, a number of works have introduced hybrid indexing\cite{chen2021spann}, leveraging the best of both indexing schemes while avoiding their limitations.

\noindent  \textbf{Generative modeling.}
Deep autoregressive networks are generative sequential models that assume a product rule for factoring the joint likelihood and model each conditional distribution through a neural network. AR models have shown powerful progress
in generative tasks across multiple domains such as image\cite{chen2020generative,yu2022scaling}, text\cite{ radford2019language,yang2019xlnet}, audio\cite{dhariwal2020jukebox,chung2019unsupervised}, and video\cite{wu2022nuwa,weissenborn2019scaling}.
The key component involves linearizing data into a sequence of symbols with notable works such as VQ-VAE\cite{van2017neural}, RQ-VAE\cite{lee2022autoregressive}.
Recently, a number of works\cite{tay2022transformer, wang2022neural, de2020autoregressive} further explored AR model to empower entity retrieval and document retrieval.
Most related to our work are NCI\cite{wang2022neural} and DSI\cite{tay2022transformer}, which are concerned with document retrieval.
However, these approaches utilize hierarchical k-means clustering applied to document embeddings derived from a small pretrained language model to obtain document identifiers. In contrast, we put forward a novel approach that involves learning the identifier directly from semantic supervision, and showcase its effectiveness in the context of image retrieval. 
We posit that this discovery can also be advantageous for document retrieval tasks.

\noindent
\textbf{Joint learning.} There is a growing interest\cite{zhan2021jointly,jang2021self,jang2020generalized,chen2020differentiable} in the study of joint learning to improve search accuracy. 
For example, JPQ\cite{zhang2021joint} jointly learns the deep retrieval model (equivalent to an embedding model) and the embedding index (compression). 
DQN\cite{10.5555/3016387.3016390} combines PQ and embedding learning but does not optimize the PQ clusters. DPQ\cite{8953697} incorporates PQ clusters for supervised joint learning, specifically targeting image retrieval. HashGAN\cite{dizaji2018unsupervised} and BGAN\cite{song2018binary} adversarially learn binary hash codes jointly with the embedding model in an unsupervised way.
However, all these prior works focus on end-to-end training of encoding and compression together while overlooking the indexing part (meaning non-exhaustive search), which is challenging to formulate as differentiable. They leverage quantization (compression index) for distance approximation but still rely on linear scan in their experiments. In contrast, our approach reframes retrieval as a completely novel end-to-end task, eliminating explicit distinctions between embedding, compression, and indexing.


\section{Method}


\subsection{Semantic Image Tokenizer}
As Transformer becomes the ubiquitous architecture in computer vision, it has emerged many successful image tokenizers such as VQ-VAE\cite{van2017neural,ramesh2021zero,gafni2022make,yu2021vector}, RQ-VAE\cite{lee2022autoregressive} and so on. 
Despite its success in image generation, they may not be well-suited for the retrieval task for several reasons. First, the process of decoding latent codes to reconstruct raw images is essential for generating images in synthesis tasks but is redundant for retrieval tasks. Second, the autoregressive model's inference speed is heavily influenced by the length of the code sequence, where a shorter sequence is vital for efficient searching in our context. Presently, code sequences tend to be lengthy (for example, an $8\times 8$ feature map with a depth of 4 in RQ-VAE leads to a sequence of 256), underscoring the need to significantly reduce sequence length. Moreover, retrieval tasks require the embedding of semantic information within the latent codes. However, the current approach to learning latent codes through image reconstruction loss primarily captures low-level details, such as inconsequential local nuances and noise, which may not be beneficial for retrieval.

Building on the insights mentioned earlier, we suggest investigating the global feature extracted from the class token, rather than relying on the default spatial tokens. This approach offers a substantial reduction in sequence length, reducing it from 64 tokens to a single token. Moreover, as a result of this approach, the class token inherently contains a condensed, high-level semantic representation.
Let $\mathbf{f}_{cls}$ denote the $d$-dimensional feature vector outputted from the class token, which is taken as the image representation. We adopt residual quantization (RQ) or stacked composite quantization to approximate this feature.
Suppose there are $M$ codebooks with each containing $L$ elements, $\mathcal{C}_m=\{\mathbf{c}_{m1},\cdots, \mathbf{c}_{mL}\}$,
RQ recursively maps the embedding $\mathbf{f}_{cls}$ to a sequentially ordered $M$ codes, $\mathbf{f}_{cls} \rightarrow \{l_1, l_2,\cdots,l_M\} \in [\mathbb{L}]^{M}$. Let $\mathbf{r}_0 = \mathbf{f}_{cls}$, we have
\begin{align}
	l_m &= {\arg \min}_{l \in [\mathbb{L}]} \|\mathbf{r}_{m-1} - \mathbf{c}_{ml}\|_2^2, \\
	\mathbf{r}_m & = \mathbf{r}_{m-1} - \mathbf{c}_{ml_m}, ~ m = 1,2,\cdots,M.
\end{align}
The process of sequentially generating discrete codes is inherently compatible with sequential autoregressive generation. This alignment helps alleviate the optimization challenges associated with modeling the relationships within identifiers. A figure illustrating the image tokenizer can be seen in the supplementary.


To further inject semantic prior, we propose to train the network under classification loss over both the original embeddings as well as the reconstructed embeddings. 
In particular, we consider a series of reconstruction levels denoted as
$\hat{\mathbf{f}}_{cls}^{\le m} = \sum_{i=1}^m \mathbf{c}_{il_i}, m=1,2,\cdots,M$.
Each prefix code thus encodes semantic information to a certain degree.
Adding up the $M$ levels of partial reconstruction error,
the complete objective function is then formulated as,
\begin{align}
	&\mathcal{L} = \mathcal{L}_{cls} (\mathbf{f}_{cls})  + \lambda_1 \sum_{m=1}^M \mathcal{L}_{cls}(\hat{\mathbf{f}}_{cls}^{\le m})  + \lambda_2 \sum_{m=1}^M \|\mathbf{r}_m\|_2^2, \\
	&\mathbf{r}_m  = \mathbf{f}_{cls}- \text{sg} [\hat{\mathbf{f}}_{cls}^{\le m}], ~ m=1,2,\cdots,M,
	\label{eqn:rqvit}
\end{align}
where $\mathcal{L}_{cls}$ is the softmax cross-entropy loss and $\text{sg}[\cdot]$ is the stop gradient operator. 
During training, we adopt alternative optimization to update the codebook and the network.
For computing the gradient of $\mathcal{L}_{cls}(\hat{\mathbf{f}}_{cls}^{\le m})$, we follow the straight-through estimator\cite{bengio2013estimating} as in\cite{van2017neural}.
After optimization, 
we hope that images with similar classes have close codes. 
In the experiments, we present comparison with other discrete identifiers including random codes and codes from hierarchical k-means algorithm as well as from RQ-VAE.

\subsection{Encoder-Decoder for Autoregressive Retrieval}
\begin{figure*}[t]
	\vskip 0.1in
	\begin{center}
		\centerline{\includegraphics[width=1\columnwidth]{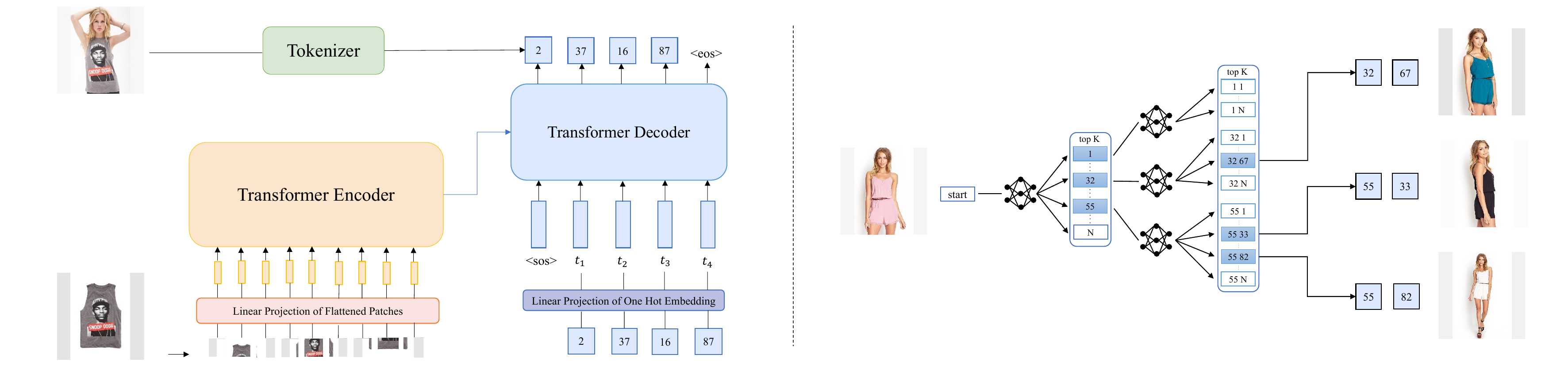}}
		\caption{The left of the figure illustrates the training pipline of encoder-decoder architecture for autoregressive retrieval. The training objective is to autoregressively predict the identifier of the query's nearest neighbor image. The right of the figure illustrates the procedure of beam search with code length is 2 and beam size K is 3.}
		\label{fig:framework}
	\end{center}
	\vskip -0.5in
\end{figure*}

Once we have established a robust discrete latent structure equipped with semantic prior information, our next step is to train
a powerful autoregressive sequence-to-sequence model solely on these discrete random variables without referring their visual content. Our encoder-decoder architecture decouples the input embedding from the generation of discrete codes. 
The training process involves image pairs $(x_1, x_2)$, where $x_2$ is the nearest neighbor of $x_1$. Our model's objective is to predict the identifiers of $x_2$ when given $x_1$ as input. This setup allows the model to learn the semantic relationships between images in the dataset. The left of \cref{fig:framework} provides a concise view of our training pipeline. After training,
the model begins by taking a query image as input to obtain the query embedding, which is decoded to the discrete codes.
It is worth noting that the yielded discrete codes represent the query's nearest neighbor images within the database rather than for the query.

To be specific, let the encoder be denoted as $\mathbb{E}$ based on the ViT base architecture and the decoder be $\mathbb{D}$, a standard Transformer that includes causal self-attention, cross-attention and MLP layers. 
We leverage the spatial tokens outputted from the encoder as the input embedding, $\mathbf{e} = \mathbb{E}(x_1)$, which is injected into the decoder through cross attention. Our training objective involves predicting the next token in the image identifier sequence. Specifically, we aim to maximize the probability of the $i$-th token of the image identifier given the input embedding and the previously predicted tokens, $p(l_i|x_1,l_1,\cdots,l_{i-1},\theta)$, where $\theta$ denotes the parameters of both $\mathbb{D}$ and $\mathbb{E}$, and $l_1,l_2,\cdots,l_M$ are the $M$ tokens that make up the image identifier for $x_2$, generated by the image tokenizer. By maximizing the probability of each token, we effectively maximize the likelihood of generating the image identifier of an image,
\begin{align}
	p(l_1,\cdots,l_M|x_1,\theta) = \prod_{m=1}^M p(l_i|x_1,l_1,\cdots,l_{m-1},\theta).
\end{align}
We apply a softmax cross-entropy loss on a vocabulary of $M$ discrete tokens during training. 

\subsection{Beam Search}
During inference, given a query image $\mathbf{q}$, we first calculate the query embedding processed by the encoder $\mathbb{E}$ and then generate discrete codes through the decoder $\mathbb{D}$ based on the query embedding in an autoregressive manner. These discrete codes represent an image that is considered as the nearest neighbor to the query image. This illustrates the process of Top-1 retrieval by generating the sequence codes one by one according to the argmax probability.
To perform Top-K retrieval, our model utilize beam search integrated into the generating process, allowing us to find the top-$K$ images that are the closest matches.



As illustrated in right of \cref{fig:framework}, this procedure begins by defining a beam size, denoted as K in our case. Initially, the beam comprises the start-of-sequence token, and the algorithm extends the current set of candidate sequences by considering the next potential tokens in the sequence. For each candidate sequence, the algorithm outputs probabilities for multiple potential next tokens and adds them to the beam. The expanded set of sequences is then scored based on the probabilities assigned by the model, which are computed as the product of the probabilities associated with their individual elements. The top-k sequences with the highest scores are retained, while the others are discarded. This selection process, which can be implemented with a priority queue in linear time with the candidate list size, is repeated until the last token is decoded. 

It's important to note that not all generated identifiers are necessarily valid, meaning that an identifier belonging to the set $[\mathbb{L}]^M$ may not correspond to any images within the retrieval database. Therefore, we must validate the generated image identifier at each step, which can be a time-consuming process. However, we address this challenge by constructing a prefix tree containing valid codes.
In the case of beam search with a prefix tree, the search process is constrained to only consider valid next tokens, as determined by the prefix tree, rather than all the possible next tokens. This constraint ensures that the search process results in valid images,
largely enhancing the efficiency of the retrieval process.


\noindent
 \textbf{Beam search vs. ANN search.}
Indeed, beam search can be regarded as a tree-based manner to perform approximate nearest neighbor (ANN) search.
However, compared to prior methods, there is a significant difference 
in how they calculate the score to choose the current node.
Traditional methods usually rely on calculating the distance between the query feature and the node's associated feature, using a predefined distance metric to determine the score. Conversely, in beam search, the score or likelihood is produced by a differentiable neural network conditioned on the query. This distinction allows the retrieval process to be refined end-to-end.

\section{Experiments}
\begin{table*}[t]
\caption{Precision comparison with different baselines, for which we consider linear scan search, Faiss IVF search and SPANN search. (repro) denotes the model reproduced by ourselves to ensure the same data process and comparable model size for fair comparison.  Our model adopts beam search for retrieval, achieving significant improvement and performing even better than linear scan search.}
\begin{center}
\vskip -0.2in
\begin{tabular}{l|cccc|cccc|cccc}
\toprule
 \multirow{2}{*}{Model} & \multicolumn{4}{c|}{In-shop}  & \multicolumn{4}{c|}{CUB200} & \multicolumn{4}{c}{Cars196}\\
\cline{2-13}
 & 1& 10&20&30&1&2&4&8& 1 & 2 & 4 & 8\\
 \hline
\rowcolor{mygray} \multicolumn{13}{l}{\textit{\textbf{Linear scan search}}} \\
\rowcolor{mygray} Res101-Img & 30.7 & 10.2 & 7.1 & 5.8 &  46.8 & 43.6 & 39.9 & 34.9 & 25.9 & 22.0 & 18.5 & 15.4\\
 \rowcolor{mygray} CLIP & 57.5 & 22.8 & 16.6 & 14.1 &  66.0 & 63.5 & 59.4 & 53.8 & 70.8 & 67.8 & 63.3 & 57.2\\
 \rowcolor{mygray} CGD$_{\text{(repro)}}$ & 83.2 & 47.8 & 40.2 & 37.0  & 76.7 & 75.5 & 73.7 & 71.4 & 87.1 & 86.1 & 84.6 & 82.6\\
  \rowcolor{mygray} IRT$_{\text{R}}$$_{\text{(repro)}}$ & 92.7 & 59.6 & 51.1 & 47.6  & 79.3 & 77.7 & 75.0 & 71.4 & 75.6 & 73.1 & 68.3 & 61.7\\
  \rowcolor{mygray} FT-CLIP & 91.4 & 66.8 & 58.9 & 55.4  & 79.2 & 77.6 & 76.0 & 73.2 & 88.4 & 87.7 & 87.1 & 85.8\\
 \hline
 \multicolumn{13}{l}{\textit{\textbf{Faiss IVF PQ search}}} \\
  CGD$_{\text{(repro)}}$ & 60.4 & 30.5 & 24.5 & 22.0 & 71.6 & 70.8 & 69.9 & 68.7 & 84.8 & 84.4 & 84.1 & 83.3\\
  IRT$_{\text{R}}$$_{\text{(repro)}}$ & 68.6 & 35.7 & 29.3 & 26.6 & 68.9 & 67.6 & 66.2 & 63.4 & 59.1 & 57.5 & 54.7 & 51.7 \\
  FT-CLIP& 63.7 & 37.0 & 30.7 & 28.0  & 72.6 & 72.1 & 71.2 &69.7 & 86.5 & 86.3 & 86.2 & 86.0\\
  \hline
 \multicolumn{13}{l}{\textit{\textbf{ScaNN search}}} \\
 CGD$_{\text{(repro)}}$ & 83.0 & 47.7 & 40.3 & 37.2 & 76.7 & 75.2 &73.8 & 71.4 & 87.1 & 86.1 & 84.5 & 82.6 \\
 IRT$_{\text{R}}$$_{\text{(repro)}}$ & 92.0 & 58.2 & 50.0 & 46.6 & 79.3 & 77.7 & 75.1 & 71.4 & 75.4 & 72.8 & 68.1 & 61.6\\
 FT-CLIP& 90.4 & 64.6 & 56.9 & 53.5 & 79.2 & 77.5 & 76.0 & 73.2 & 88.3 & 87.7 & 87.1 & 85.8 \\
 \hline
 \multicolumn{13}{l}{\textit{\textbf{SPANN search}}} \\
 CGD$_{\text{(repro)}}$ & 83.0 & 47.7 & 40.3 & 37.1 & 76.7 & 75.5 & 73.7 & 71.4 & 87.0 & 86.1 & 84.6 & 82.6\\
 IRT$_{\text{R}}$$_{\text{(repro)}}$ & 91.4 & 56.2 & 47.9 & 44.5  & 79.3 & 77.6 & 75.0 & 71.4 & 74.8 & 72.4 & 67.6 & 61.1\\
 FT-CLIP& 90.2 & 62.9 & 55.1 & 51.8  & 78.5 & 77.6 & 76.0 & 73.2 & 88.6 & 88.1 & 87.5 & 86.3\\
 \hline
 \multicolumn{13}{l}{\textit{\textbf{Beam search}}} \\
 IRGen (ours) & \textbf{92.4} & \textbf{87.0} & \textbf{86.6} & \textbf{86.5}  & \textbf{82.7} & \textbf{82.7} & \textbf{83.0} & \textbf{82.8} & \textbf{90.1} & \textbf{89.9} & \textbf{90.2} & \textbf{90.5}\\
 \hline
\end{tabular}
\end{center}
\vskip -0.1in

\label{tab:precision}
\vskip -0.2in
\end{table*}

We evaluate our method on common image retrieval datasets and on two large-scale datasets, ImageNet\cite{deng2009imagenet} and Places365\cite{zhou2017places}. For a detailed description of the datasets and implementation details, please refer to the supplementary.

\noindent \textbf{Baselines.}
We evaluate our model's performance in comparison to five competitive baselines:
1) ResNet-101\cite{he2016deep} trained from ImageNet dataset, denoted as Res101-Img, which is commonly used as a feature extraction tool for various tasks;
2) CLIP\cite{radford2021learning} trained on 400M image-text pairs, known for powerful zero-shot capability;
3) CGD\cite{jun2019combination}, a state-of-the-art method based on ResNet;
4) IRT\cite{el2021training}, a Transformer-based model for image retrieval and we use the best-performing model IRT$_{\text{R}}$;
5) FT-CLIP, a baseline finetuned from CLIP on the target dataset. For both CGD and IRT, we have reproduced these models to ensure consistent data processing and comparable model sizes. Specifically, we use ResNet-101 for CGD and DeiT-B for IRT. We also provide their best results from their original papers for reference.

\begin{table*}[t]
\caption{Recall comparison with different baselines, for which we consider linear scan search, Faiss IVF search and SPANN search. (repro) denotes the model reproduced by ourselves to ensure the same data process and comparable model size for fair comparison. we include the best result of CGD and IRT from their original papers for context with * denotation. Our model adopt beam search for retrieval, achieving comparable performance in most cases. }
	\small
	\begin{center}
 \vskip -0.2in
		\begin{tabular}{l|cccc|cccc|cccc}
			\toprule
			\multirow{2}{*}{Model} & \multicolumn{4}{c|}{In-shop} & \multicolumn{4}{c|}{CUB200} & \multicolumn{4}{c}{Cars196}\\
			\cline{2-13}
			& 1& 10&20&30&1&2&4&8& 1 & 2 & 4 & 8\\
			\hline
			\rowcolor{mygray} \multicolumn{13}{l}{\textit{\textbf{Linear scan search}}} \\
			\rowcolor{mygray} Res101-Img & 30.7 & 55.9 & 62.7 & 66.8 & 46.8 & 59.9 & 71.7 & 80.8 & 25.9 & 35.6 & 47 & 59.7\\
			\rowcolor{mygray} CLIP & 57.5 & 83.0 & 87.5 & 89.7  & 66.0 & 78.1 & 87.7 & 93.5 & 70.8 & 82.6 & 91.1 & 95.9\\
			\rowcolor{mygray} CGD* & 91.9 & 98.1 & 98.7 & 99.0  & 79.2 & 86.6 & 92.0 & 95.1 & 94.8 & 97.1 & 98.2 & 98.8\\
			\rowcolor{mygray} IRT$_{\text{R}}$* & 91.9 & 98.1 & 98.7 & 99.0  & 76.6 & 85.0 & 91.1 & 94.3 & - & - & - & -\\
			\rowcolor{mygray} FT-CLIP & 91.4 & 97.3 & 98.1 & 98.5 & 79.2 & 85.0 & 89.3 & 92.0 & 88.4 & 90.5 & 92.5 & 93.8\\
			\hline
			\multicolumn{13}{l}{\textit{\textbf{Faiss IVF PQ search}}} \\
			CGD$_{\text{(repro)}}$ & 60.4 & 76.0 & 77.1 & 77.4 & 71.6 & 77.4 & 81.5 & 84.2 & 84.8 & 88.0 & 89.8 & 91.0\\
			IRT$_{\text{R}}$$_{\text{(repro)}}$ & 68.6 & 79.2 & 80.0 & 80.2  & 68.9 & 77.9 & 85.0 & 89.3 & 59.1 & 70.4 & 78.2 & 83.4\\
			FT-CLIP& 63.7 & 70.7 & 71.1 & 71.2  & 72.6 & 78.0 & 82.3 & 85.2 & 86.5 & 86.9 & 87.2 & 87.5\\
			\hline
			\multicolumn{13}{l}{\textit{\textbf{ScaNN search}}} \\
			CGD$_{\text{(repro)}}$ & 83.0 & 94.8 & 96.2 & 96.7 & 76.7 & 83.5 & 88.0 & 91.8  & 87.1 & 91.7 & 94.6 & 96.6 \\
			IRT$_{\text{R}}$$_{\text{(repro)}}$ & 92.0 & \textbf{97.8} & \textbf{98.3} & \textbf{98.4} & 79.3 & 86.8 & 91.9 & 94.7 & 75.4 & 84.7 & 90.9 & 95.0 \\
			FT-CLIP & 90.4 & 95.9 & 96.6 & 96.9 & 79.2 & 85.0 & 89.2 & 92.7 & 88.3 & 90.5 & 92.4 & 93.7\\
			\hline
			\multicolumn{13}{l}{\textit{\textbf{SPANN search}}} \\
			CGD$_{\text{(repro)}}$ & 83.0 & 95.0 & 96.4 & 96.9  & 76.7 & 83.4 & 87.9 & 91.8 & 87.0 & 91.7 & \textbf{94.6} & \textbf{96.7}\\
			IRT$_{\text{R}}$$_{\text{(repro)}}$ & 91.4 & 97.2 & 97.6 & 97.7  & 79.3 & \textbf{86.8} & \textbf{91.9} & \textbf{94.7} & 74.8 & 84.3 & 90.5 & 94.7\\
			FT-CLIP & 90.2 & 95.8 & 96.7 & 97.0  & 78.5 & 85.0 & 89.4 & 92.9 & 88.6 & 90.7 & 92.5 & 94.2\\
			\hline
			\multicolumn{13}{l}{\textit{\textbf{Beam search}}} \\
			IRGen (ours) & \textbf{92.4} & 96.8 & 97.6 & 97.9 & \textbf{82.7} & 86.4 & 89.2 & 91.4 & \textbf{90.1} & \textbf{92.1} & 93.2 & 93.7\\
			\hline
		\end{tabular}
	\end{center}
	\vskip -0.1in
	
	\label{tab:recall}
	\vskip -0.2in
\end{table*}

\noindent \textbf{Search process.}
The baseline models primarily focus on effective feature learning. After training, these models are utilized to extract features for the database images. During the search process, a given query image is initially passed through the model to obtain its query feature. Subsequently, this query feature is compared with the features of database images using a specific distance metric. Following the conventions established in previous works such as\cite{radford2021learning,jun2019combination,el2021training}, we employ the cosine distance for the CLIP model and the Euclidean distance for the other baseline models.
We evaluate two search strategies: linear scan search (K-nearest neighbors or KNN) and approximate nearest neighbor search (ANN). Linear scan search is known for its accuracy but is computationally intensive. In contrast, ANN is significantly more efficient. For ANN, we explore: (i) the popular Faiss IVF PQ\cite{johnson2019billion};
(ii) the state-of-the-art memory-based algorithm ScaNN\cite{guo2020accelerating} with the default setting;
and (iii) the state-of-the-art disk-based SPANN algorithm\cite{chen2021spann}.
These evaluation strategies allow us to assess the retrieval performance of our model against a variety of search methods.

\subsection{Results}

\cref{tab:precision} presents a detailed performance comparison in terms of precision@$K$, which assesses the percentage of retrieved candidates that share the same class as the query among the top $K$ results. Our model consistently outperforms all other models, even surpassing those employing linear scan search. Notably, we achieve remarkable improvements, such as a 20.2\% boost in precision@10 on the In-shop Clothes dataset, a 6.0\% increase in precision@2 on the CUB200 dataset, and a 2.4\% gain in precision@2 on the Cars196 dataset.
Furthermore, several observations can be made: 1) Finetuned models, tailored to specific datasets, exhibit significantly better performance compared to off-the-shelf feature extractors like CLIP and ImageNet-pretrained ResNet-101.
2) Generally, models equipped with ANN algorithms perform slightly worse than their counterparts using linear scan search. However, there are exceptions, such as FT-CLIP with SPANN search on the Cars196 dataset, which demonstrates the importance of end-to-end optimization. 3) Our model consistently maintains high precision scores as $K$ increases, while other models experience a substantial drop.

\cref{tab:recall} provides a comparison of different models using the Recall@$K$ metric. Recall@$K$ measures the proportion of queries for which at least one image among the top $K$ retrieved candidates shares the same label as the query image, yielding a score of 1 if true and 0 otherwise. The table also includes the best recall results of CGD and IRT from their respective original papers for reference. It's important to note that these models may have different data preprocessing, model sizes, and additional training techniques.
Here are the key observations:
1) Our IRGen model achieves the highest Recall@$1$ score compared to all other models. However, for other recall scores, our model performs similarly or slightly worse. This discrepancy may arise from the current objective loss used in autoregressive models, which heavily optimizes for Recall@1 while giving less emphasis to other recall values. One potential solution is to incorporate the beam search process into training for joint optimization.
2) Different combinations of feature extractors and ANN algorithms exhibit significant variations across the three datasets, highlighting the challenges of achieving coordination in practical scenarios.
3) Notably, despite the high recall achieved by baselines, they often require an additional re-ranking stage to improve precision, whereas our model already attains high precision scores without the need for re-ranking.

\begin{figure*}[t]
	\centering
	(a)\includegraphics[width=.31\linewidth, clip]{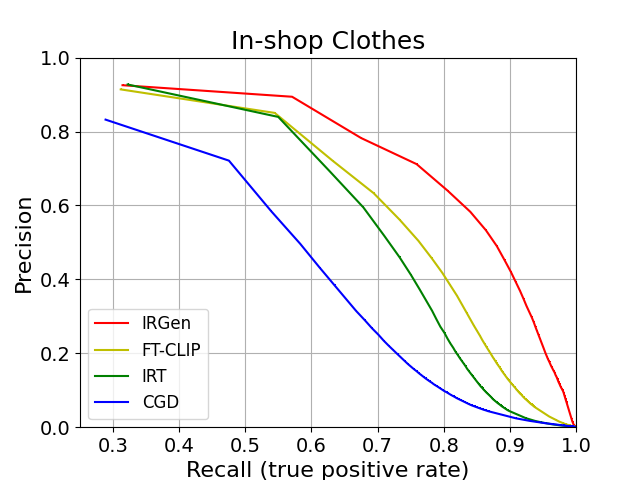}~~
	(b)\includegraphics[width=.31\linewidth, clip]{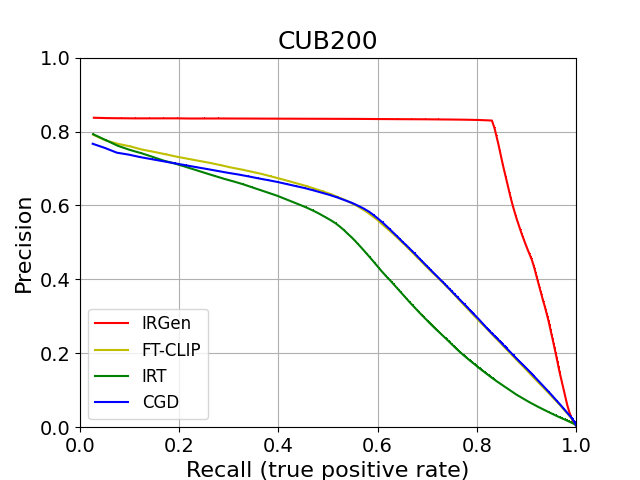}~~
	(c)\includegraphics[width=.31\linewidth, clip]{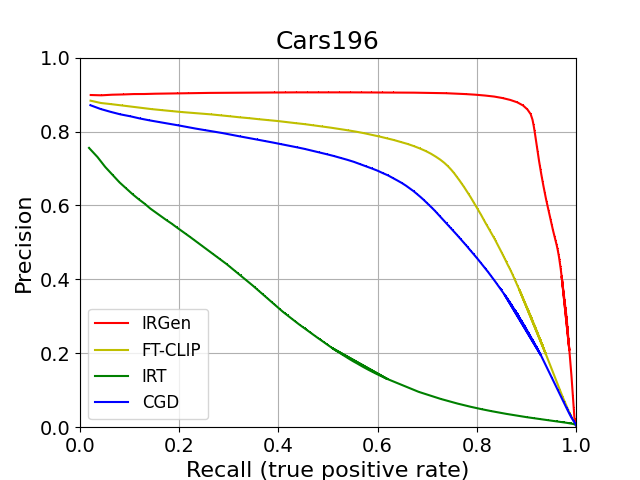}
	\caption{ Precision-Recall (TPR) curve comparison for different methods on (a) In-shop Clothes, (b) CUB200 and (c) Cars196 dataset.}
	\label{fig:pr}
	\vspace{-1em}
\end{figure*}

\begin{figure*}[t]
	\centering
	(a)\includegraphics[width=.31\linewidth, clip]{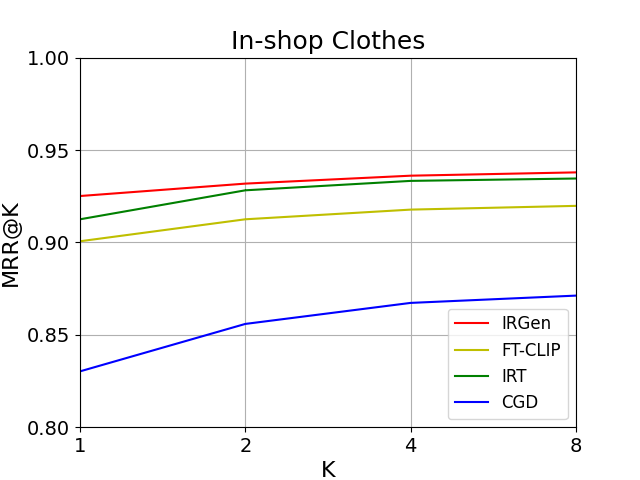}~~
	(b)\includegraphics[width=.31\linewidth, clip]{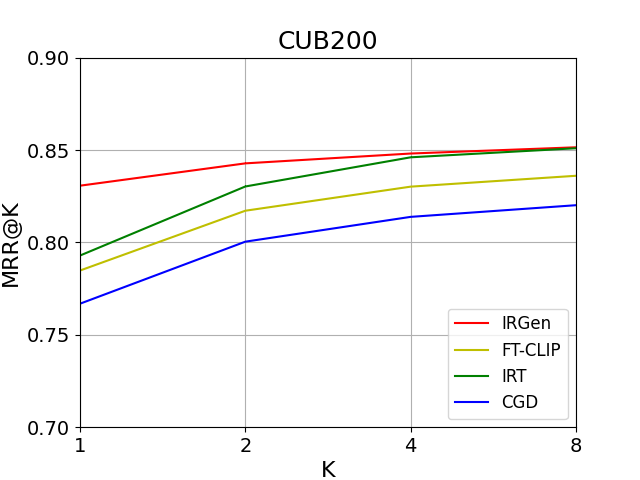}~~
	(c)\includegraphics[width=.31\linewidth, clip]{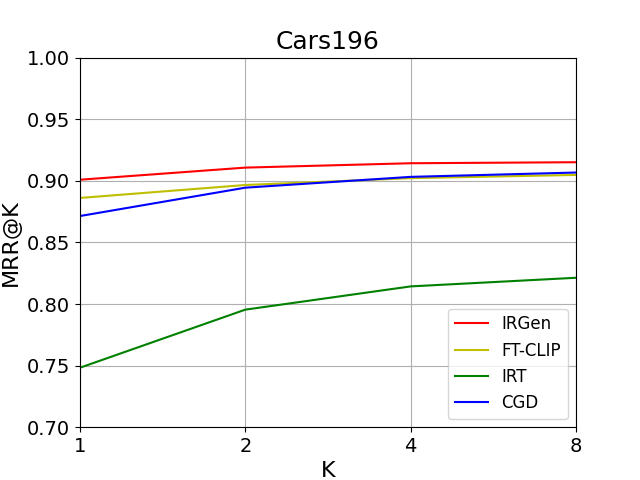}
	\caption{MRR with respect to 1,2,4,8 comparison for different methods on (a) In-shop Clothes, (b) CUB200 and (c) Cars196 dataset. }
	\label{fig:mrr}
	\vskip -0.2in
\end{figure*}

\cref{fig:pr} illustrates the precision-recall curve, where recall represents the true positive rate. Our approach, IRGen, consistently delivers outstanding performance, maintaining high precision and recall simultaneously.
In addition to precision-recall analysis, we evaluate our model using the mean reciprocal rank (MRR) metric, which measures the inverse of the rank of the first relevant item. We compute MRR for four different values: 1, 2, 4, and 8, and display the corresponding curves in \cref{fig:mrr}. The baselines employ the SPANN retrieval algorithm. 
Our IRGen model consistently outperforms the baselines across all evaluated metrics, confirming the effectiveness of our framework.
Notably, there is significant variability in the performance gap between each baseline and our model across the three datasets, highlighting the challenges and dataset-dependent nature of retrieval tasks.



\begin{table}[t]
	\centering
        \begin{minipage}[b]{0.45\textwidth}
         \caption{MAP@100 on two million-level datasets. The results of CLIP and FT-CLIP are retrieved by SPANN. }
        \centering
		\setlength\tabcolsep{4pt}
		\footnotesize
 
		\begin{tabular}{l|cc}
			\toprule
            \multirow{2}{*}{Model} & \multicolumn{2}{c}{Dataset} \\
            \cline{2-3}
            & ImageNet & Places365 \\
            \hline
            CLIP & 44.1 & 22.1 \\
            FT-CLIP & 65.5 & 30.3\\
            IRGen(Ours) & \textbf{76.0} & \textbf{44.3} \\
			\hline
		\end{tabular}

	
	\label{tab:map@100}
        \end{minipage}
    \begin{minipage}[b]{0.45\textwidth}
    \caption{MAP@1000 for 32 bits (equivalent to the length of our identifier) comparison on ImageNet dataset.  }
    \centering
    \setlength\tabcolsep{4pt}
		\footnotesize
		\begin{tabular}{l|c}
			\toprule
			Method & MAP@1000 for 32 bits\\
			\hline      
            DTQ\cite{liu2018deep} & 68.12\\
            ADSVQ\cite{zhou2020angular} & 75.21\\
            DPQ\cite{8953697} & 87.70\\
            IRGen(Ours) & \textbf{93.52}\\
			\hline
		\end{tabular}
	
	\label{tab:joinlearningonimagenet}
 \end{minipage}
\end{table}




\begin{table}[t]
\caption{Generalize to new data. We split 5\% of the training data from the ImageNet dataset for inference and remained unseen during training.}
\vskip -0.2in
\begin{center}
\setlength\tabcolsep{5pt}
\begin{footnotesize}
\vskip -0.2in
\begin{tabular}{l|ccc}
			\hline
			\multirow{2}{*}{Model} & \multicolumn{3}{c}{Precision} \\
			\cline{2-4}
			& 1 & 10 & 100 \\
			\hline
			FT-CLIP + Linear Scan & 70.6 & 65.0  & 55.6\\
			IRGen (Ours) & \textbf{77.0} & \textbf{77.9} & \textbf{77.4}\\
			\hline
		\end{tabular}
\end{footnotesize}
\end{center}

\label{tab:newdata}
\vskip -0.3in
\end{table}

\begin{figure}[t]
	\centering
	\includegraphics[width=.48\linewidth, clip]{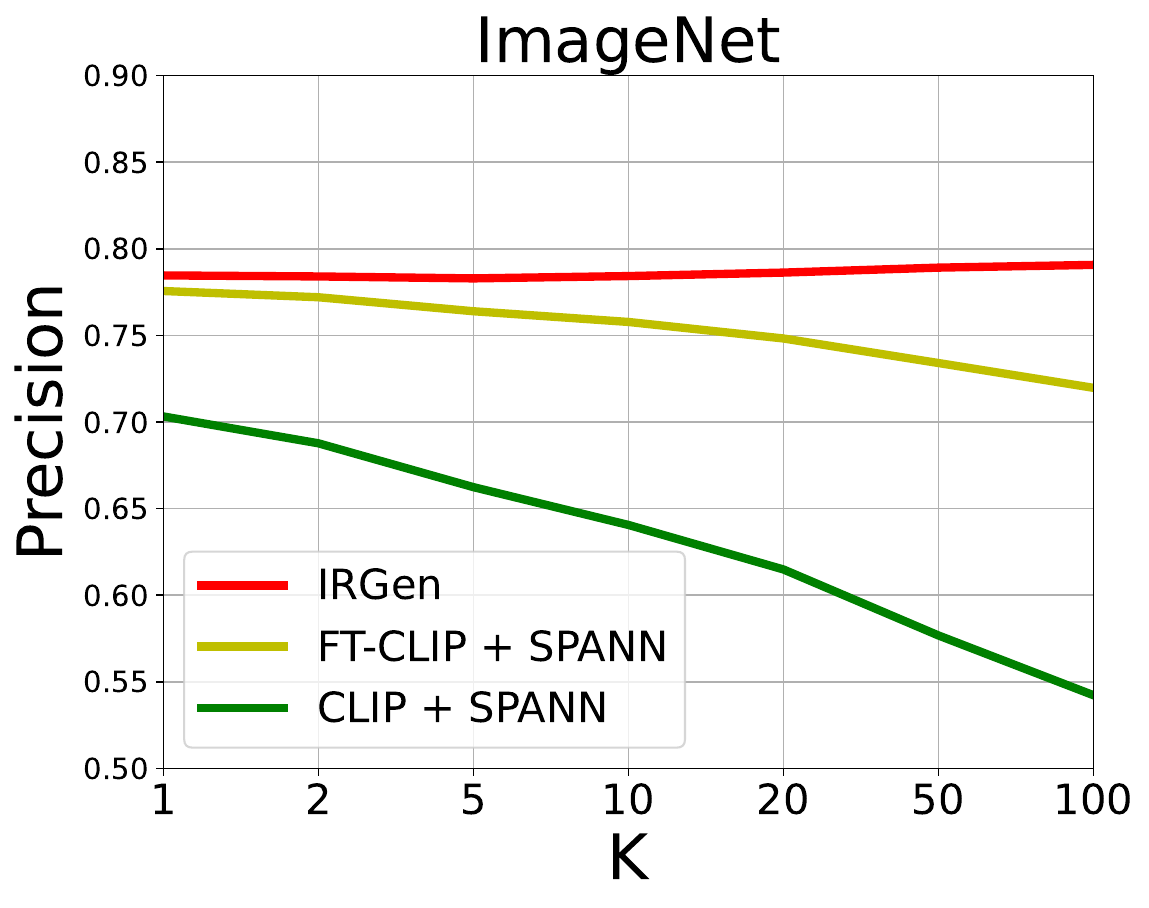}~
	\includegraphics[width=.48\linewidth, clip]{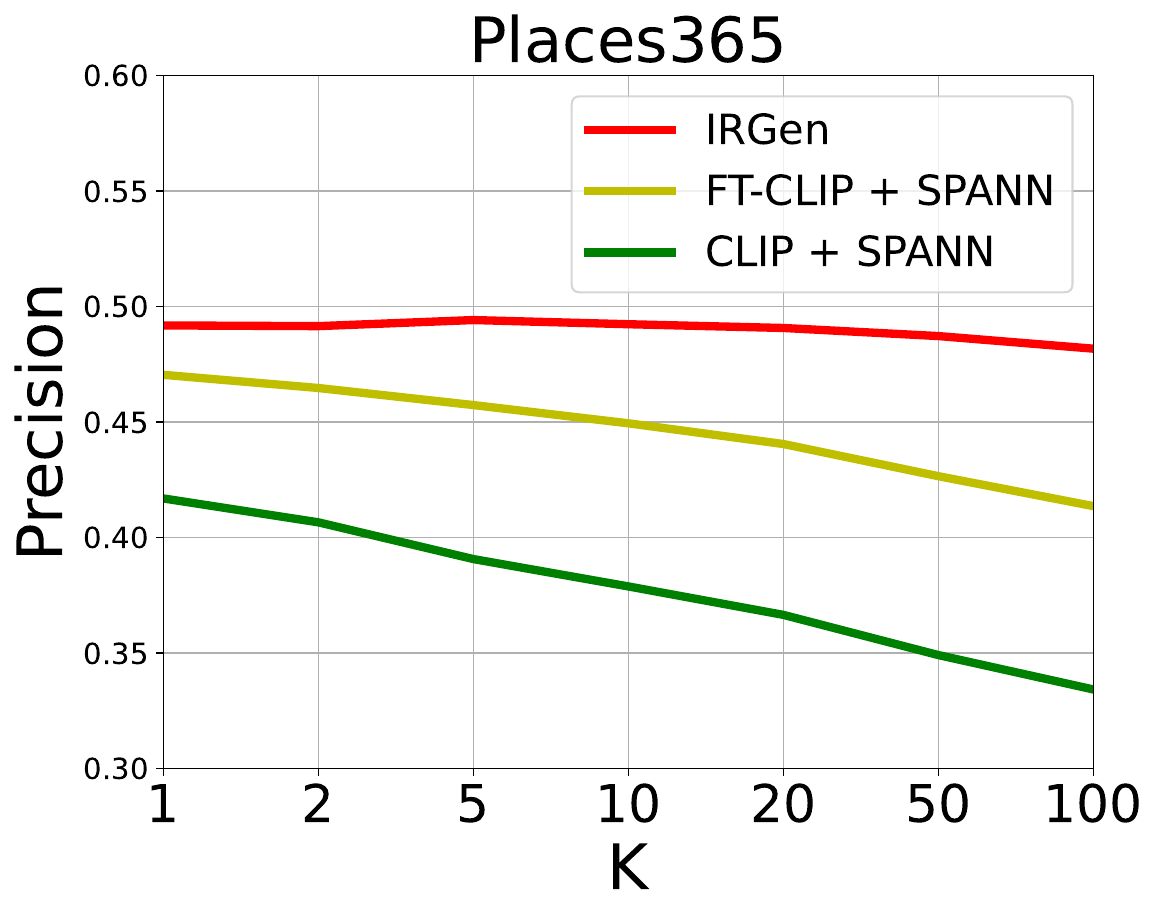}
	\vskip -0.2in
	\caption{ Precision comparison on large scale datasets: ImageNet and Places365.}
	\label{fig:million}

\end{figure}

\noindent \textbf{Results on million-level datasets.}
We further experiment our approach with ImageNet dataset\cite{deng2009imagenet} that contains 1,281,167 images and Places365-Standard\cite{zhou2017places} containing about 1.8$M$ images from 365 scene categories.
We compare with the strong baselines including CLIP model as well as FT-CLIP model finetuned based on CLIP model.  The  comparison is reported in \cref{fig:million} and \cref{tab:map@100}, focusing on precision@$K$ and MAP@100. 
Our IRGen model consistently outperforms the baselines, achieving the best results in terms of precision@$K$ and MAP@100.
The precision values for our model remain consistently high as $K$ increases, while the baselines experience noticeable performance degradation.
These results confirm the effectiveness of our model in handling large-scale datasets like ImageNet, where it maintains high precision across varying values of $K$ and outperforms the baseline models.

\noindent \textbf{Comparison to supervised deep quantization methods.}
We make the comparison with recent supervised deep quantization methods on ImageNet, including the state-of-the-art methods ADSVQ\cite{zhou2020angular} and DPQ\cite{8953697}. Notably, these methods exclusively utilize 100 classes from ImageNet, as indicated in their original papers, making it a substantially easier case. According to the original paper, DPQ\cite{8953697} employed ResNet-50 (25.56M parameters), ADSVQ\cite{zhou2020angular} utilized AlexNet (60.3M parameters), and DTQ\cite{liu2018deep} also used AlexNet. To ensure fair comparison, we employ a smaller network, ViT-S (22.1M parameters) and rerun over 100 classes. \cref{tab:joinlearningonimagenet} illustrates the superiority of our approach.


\subsection{Ablations}

\noindent \textbf{The effect of identifiers.}
In our study of image identifiers, we compared four different approaches: (1) assigning random identifiers to images, (2) hierarchical k-means (HKM), (3) Product Quantization (PQ) and (4) using the image tokenizer RQ-VAE\cite{bevilacqua2022autoregressive}. The results of this comparison are summarized in \cref{tab:identifier}.
The random assignment of identifiers to images yielded expectedly lower performance. This performance gap can be attributed to the fact that models with random identifiers need to learn not only the interaction between queries and image identifiers but also allocate capacity to learn relationships within the identifiers themselves.
On the other hand, HKM showed superior performance compared to random assignment, underscoring the significance of semantic identifiers. However, our proposed semantic image identifiers demonstrated a clear improvement over HKM, highlighting their effectiveness in enhancing retrieval performance.
PQ codes perform worse than the random assigned identifiers. This underperformance may stem from the model's challenges in constructing semantic information from PQ codes, which might be incompatible with autoregressive prediction mechanism. This highlights the significance of our sequence order in autoregressive training process.
In contrast, the performance of RQ-VAE significantly trailed behind our model, with a performance less than 10 percent. We attribute this difference to the sequence length in RQ-VAE which is too long for the model to effectively capture relationships within the identifiers. 

\noindent \textbf{Generalize to new data.}
Addressing the inclusion of fresh data holds particular significance, especially in the context of search scenarios. To assess this capacity, we conduct an experiment where we intentionally withheld 5\% of the training data from the ImageNet dataset during the training phase and introduced it during inference, all without updating the existing codebook and AR model. In this experiment, we compare our model with the formidable baseline FT-CLIP, which is equipped with a linear scan search.
The results, as displayed in \cref{tab:newdata}, reveal that our model maintains superior performance even when confronted with new data. This observation highlights our model's ability to effectively generalize to previously unseen data.
In scenarios where more new data come in with vastly different semantic structure, leading to the distribution of gallery data change drastically, all data-dependent component such as embedding model as well as joint learning methods including IRGen, have to re-train the model and re-construct the data index. Essentially, generalization (or dynamic updates) only supports a small number of new data that do not substantially alter the distribution which we have demonstrated in our case. In future we may extend IRGen to an adaptive structure capable of capturing the evolving distribution of new data during training.

\begin{table}[t]
\caption{Ablation study on the image identifier (T=length).}

\begin{center}
\setlength\tabcolsep{3.5pt}
\begin{footnotesize}
\vskip -0.2in
\begin{tabular}{l|c|cccc|cccc}
\toprule
 \multirow{2}{*}{Identifier} &  \multirow{2}{*}{T} & \multicolumn{4}{c|}{Precision} & \multicolumn{4}{c}{Recall} \\
\cline{3-10}
 & & 1& 10&20&30& 1 & 10&20&30\\
 \hline
Random & 4 & 87.6 & 75.4 & 70.8 & 68.3 & 87.6 & 95.1 & 96.0 & 96.1\\
 HKM$_{100}$ & 4 & 88.2 & 80.0 & 78.2 & 77.3 & 87.2 & 93.1 & 94.3 & 95.0\\
 HKM$_{200}$ & 3 & 89.0 & 81.6 & 79.8 & 79.0 & 89.0 & 93.9 & 94.9 & 95.8\\
 HKM$_{500}$ & 2 & 89.5 & 81.7 & 79.8 & 78.9 & 89.5 & 95.3 & 96.5 & 97.0\\
 PQ &  4 & 73.8 & 66.1 & 61.8 & 59.2 & 73.8 & 90.2 & 93.3 & 94.5 \\
 Ours & 4 &\textbf{92.1} & \textbf{89.7} & \textbf{89.5} & \textbf{89.5} & \textbf{92.1} & \textbf{96.8} & \textbf{97.5} & \textbf{97.7}\\
 \hline
\end{tabular}
\end{footnotesize}
\end{center}
\vspace{-1em}
\label{tab:identifier}
\vskip -0.2in
\end{table}


\begin{table}[t]

\caption{Comprehensive comparison of storage requirement and inference cost.}
\vspace{0em}
\label{tab:cost}
\vskip -0.3in
\begin{center}
\setlength\tabcolsep{3pt}
\vskip -0.3in
\begin{tabular}{l|cc}

\toprule
Method & Model(Params) & Database storage\\
\hline
IRGen & ViT-base encoder(86M)+24-layer decoder(307M) & 4.9MB$:$ 4 bytes/image\\
FT-CLIP & ViT-base(86M) & 3.67GB$:$ 768x4 bytes/image\\
 \hline
\end{tabular}
\vskip -1.5in

\end{center}

\end{table}

\begin{figure}[t]
\vskip -0.2in
\begin{center}
\centerline{\includegraphics[width=.6\columnwidth]{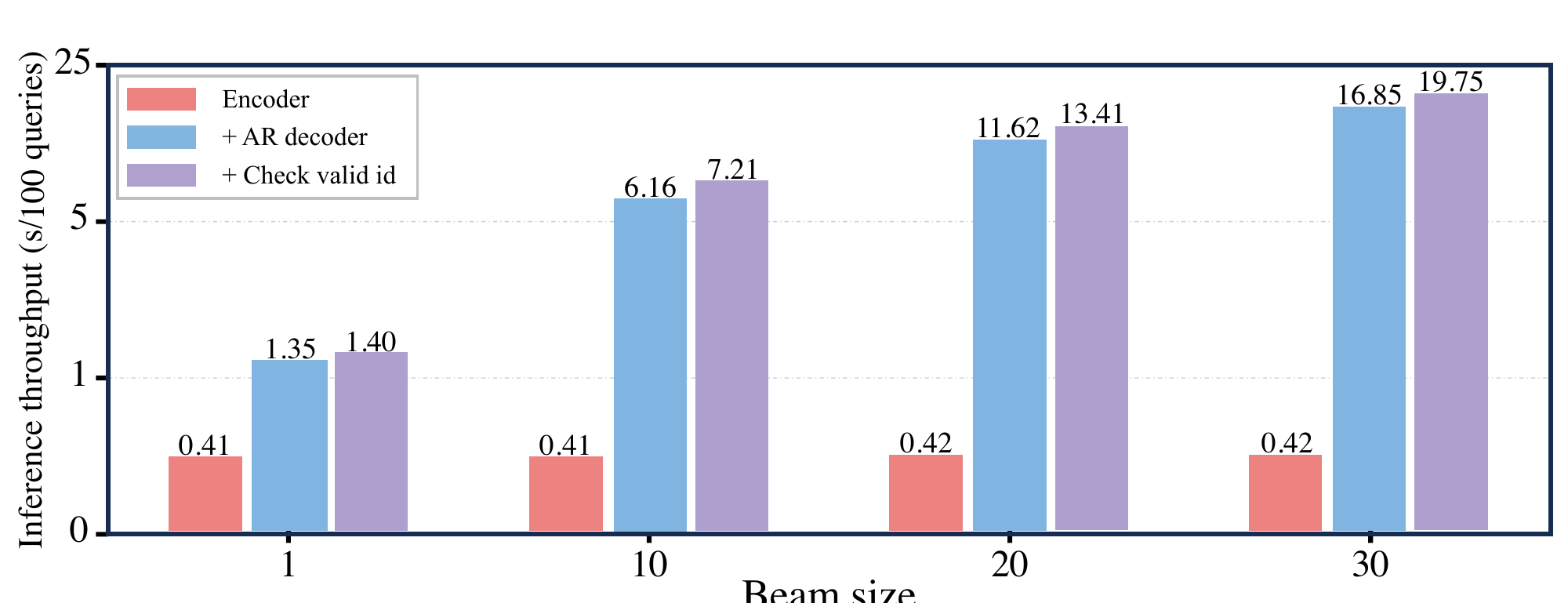}}
\vspace{-1em}
\caption{Illustrating the search speed using beam search.}
\label{fig:throughput}
\end{center}
\vskip -0.4in
\end{figure}

\noindent \textbf{Storage requirement and Inference throuput.}
We first provide a comprehensive comparison of model capacity on the ImageNet dataset and the comparison is shown in the \cref{tab:cost}. It is crucial to emphasize that a retrieval system's storage considerations extend beyond just the model; it must also store features for all database vectors, constituting a substantial portion of storage requirements, especially in the context of a large database. While compression can alleviate this concern, the performance is severely compromised due to quantization. 
Our GPU inference cost, tested on a Tesla P100 PCIe 16GB GPU, is 0.037s/image, which is near-real-time, considering practical retrieval times at the order of 10 milliseconds. Moreover, our approach achieves 76\% MAP@100, Moreover, our approach achieves 76\% MAP@100, significantly surpassing the FT-CLIP with SPANN result of 65.5\%, and is competitive with FT-CLIP utilizing linear scan, which achieves 77\%. Notably, however, the FT-CLIP with linear scan method demands 0.14 seconds per image and considerable memory usage.

We further report additional inference throughput on Inshop dataset for 100 queries, with beam sizes set at 1, 10, 20, and 30 for comparison.
This analysis is conducted on an NVIDIA V100-16G GPU and we break down the time cost of each component during autoregressive retrieval.
As shown in \cref{fig:throughput}, 
the encoder is quite fast, while the autoregressive decoder becomes the major bottleneck, especially as the beam size increases. Additional time is consumed for checking the validity of predictions, as it's possible that the predicted identifier may not exist in the database. Overall, the time cost is within an acceptable range. For instance, it takes approximately 0.07 seconds (with a beam size of 10) or 0.19 seconds (with a beam size of 30) per query.

For inference speed, we anticipate that the prevalent reliance on GPUs  will swiftly lead to newer GPU generations with increased processing power, more efficient memory management, and improved hardware support, enhancing the speed of our autoregressive retrieval. Additionally, our model, being GPT-based, can leverage rapidly developed acceleration techniques designed for GPT.

\section{Conclusion}
In this paper, we delve into the realm of generative modeling to enable end-to-end image retrieval, a process that directly connects a query image to its closest match. With the introduction of semantic image tokenizer, we've demonstrated that our model excels at achieving remarkable precision without compromising recall. Through extensive ablation studies and evaluations, we've underscored the superior performance of our approach. We believe that this innovative approach to generative modeling in image retrieval not only pushes the boundaries of this field but also holds potential for broader applications.

\noindent \textbf{Limitations.}
While our model has shown significant performance improvements, it's important to acknowledge its limitations.
First,
though our model has demonstrated scalability to million-scale datasets, dealing with billion-scale datasets is a complex challenge. It may necessitate even larger models with higher capacity, potentially impacting inference speed. Striking a balance between model capacity and speed is an area that warrants exploration for efficient and effective billion-scale search.
Second
Training large autoregressive models requires substantial computational resources, which raises environmental concerns. Research efforts to enable efficient training, such as fast fine-tuning of pretrained models, are crucial to mitigate energy consumption and environmental impact.

\section*{Acknowledgements}
This work is supported by Fundamental Research Funds for the Central Universities (2243100004).

%
%
\bibliographystyle{splncs04}
\bibliography{main}

\clearpage

\appendix

    \section{Image Tokenizer Detail}
\begin{figure*}
		\centering
		\includegraphics[width=.85\linewidth, clip]{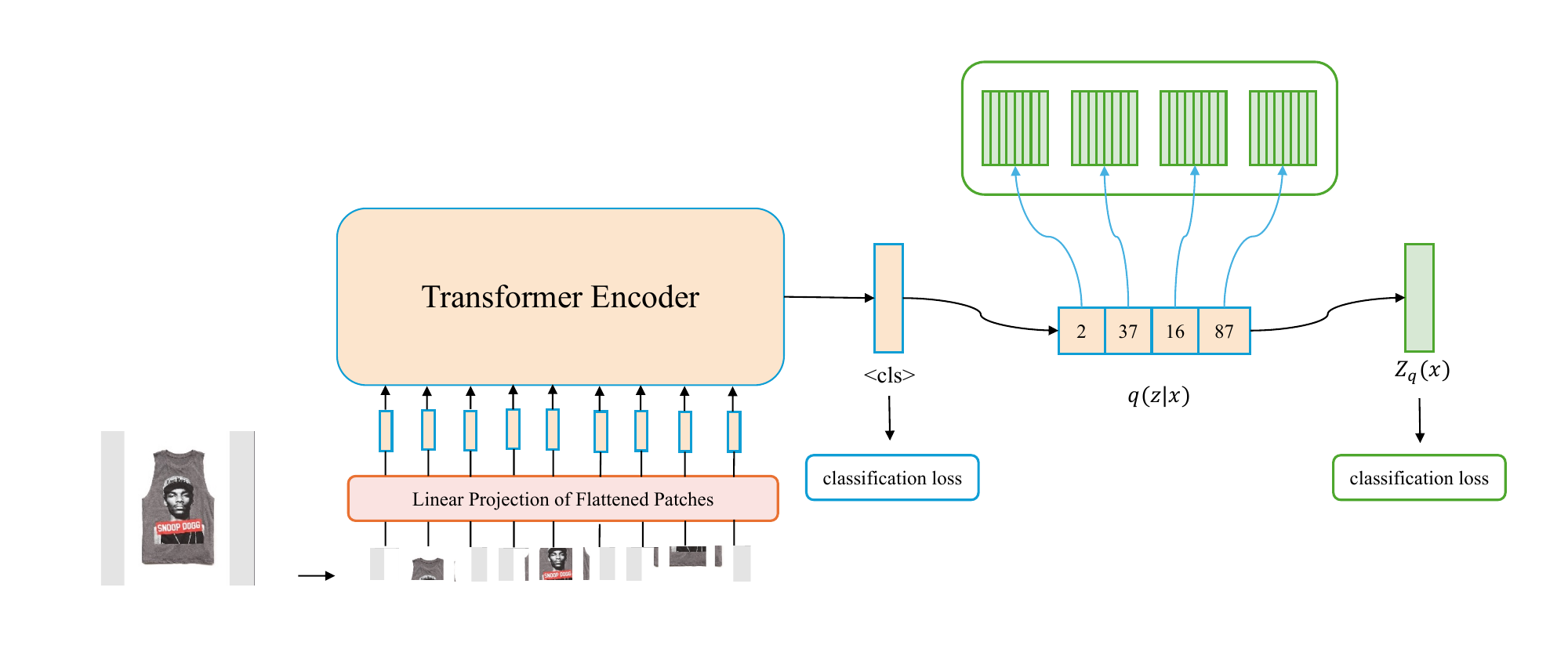}
		\caption{The framework of the image tokenizer.}
		\label{fig:tokenizer}
	\end{figure*}

    \cref{fig:tokenizer} shows the architecture of our image tokenizer, which is an encoder only network where we enforce classification loss over feature embeddings. We indeed utilize the feature from the ”class token” to perform residual quantization. 

	\section{Dataset Detail}

	\noindent \textbf{In-shop Clothes} retrieval dataset~\cite{liu2016deepfashion} is a large subset of DeepFashion with large pose and scale variations.  This dataset consists of a training set containing 25,882 images with 3997 classes, a gallery set containing 12,612 images with 3985 classes and a query set containing 14,218 images with 
	3985 classes. The goal is to retrieve the same clothes from the gallery set given a fashion image from the query set. We use both the training set and the gallery set for training in our experiments.
	
	\noindent \textbf{CUB200}~\cite{wah2011caltech} is a fine-grained dataset containing 11,788 images with 200 classes belong to birds. There are 5,994 images for training and 5,794 images for testing.
	
	\noindent \textbf{Cars196}~\cite{krause20133d} is also a fine-grained dataset about cars. It contains 16,185 images with 196 car classes, which is split into 8,144 images for training and 8,041 images for testing.
	
	\noindent \textbf{ImageNet} dataset~\cite{deng2009imagenet}  contains 1,281,167 images 
	for training and 50,000 validation images for testing, in which we randomly sample 5,000 images as queries to speed up the evaluation process.
	
	\noindent \textbf{Places365-Standard}~\cite{zhou2017places} includes about 1.8$M$ images from 365 scene categories, where there are at most 5000 images per category.

	\section{Implementation Detail}
 We adopt ViT-B for encoder and similar architecture for decoder (12 transformer decoder block with dimension 768).
	The input image is of resolution $224\times 224$ and is partitioned to $14 \times 14$ patches with each patch sized $16 \times 16$. 
	Intuitively, a warm initialization of encoder should largely stable the training process.
	We thus warm-start the model with encoder initialized by the pretrained CLIP model~\cite{radford2021learning}. 
	We randomly initialize the remaining fully connected layer and the decoder. 
	The semantic image tokenizer is trained with a batch size of 128 on 8 V100 GPUs with 32G memory per card for 200 epochs.  We adopt an AdamW optimizer~\cite{loshchilov2017decoupled} with betas as $(0.9, 0.96)$ and weight decay as $0.05$. We use cosine learning rate scheduling. Note that we set the initial learning rate as $5e-4$ for the FC layers. The learning rate of the encoder is set as one percentage of the learning rate of FC layers. We train our models with 20 warming-up epochs and the initial learning rate is $5e-7$.
	For training autoregressive model, we select similar image pairs $(x_1,x_2)$. Since current retrieval datasets are usually labeled with class information, we randomly sample an image  $x_2$ which shares the same class with  $x_1$ as the nearest neighbor.
	For autoregressive model, we use batch size of 64 on 8 V100 GPUs with 32G memory per card for 200 epochs. The optimizer and the scheduler are same as the semantic image tokenizer mentioned above. The initial learning rate is $4e-5$ for the decoder and the learning rate for encoder is always one percentage of that for decoder. The hyperparameter for quantization is set to $M=4$ and $L=256$ for fast inference. For ImageNet and Places365, the experimental settings are the same as before except that we enlarge the layer of decoder to 24 to increase the capacity for AR modeling. 
	
	\section{Ablation Study}
	\noindent \textbf{The effect of sequence length.} We further investigate the length of identifier in our image tokenizer. We experiment different lengths and report the results in \cref{tab:length}.
	We can see that 
	if the length of the identifier is too small (for example 2),
	the model gets inferior performance. 
	As with the length gets longer to 4 or 6, the model gets better performance. At last the performance drops a little bit if the length is too long (8). We think 4-6 would be a good choice in most cases and we simply use 4 in all our experiments.

	\begin{table}[h]
  \caption{Ablation study on the sequence length T.}
		\begin{center}
			\begin{small}
   \vskip -0.2in
				\begin{tabular}{c|cccc|cccc}
					\toprule
					\multirow{2}{*}{T} & \multicolumn{4}{c|}{Precision} & \multicolumn{4}{c}{Recall} \\
					\cline{2-9}
					& 1& 10&20&30& 1 & 10&20&30\\
					\hline
					2 & 72.1 & 69.6 & 68.9 & 68.6 & 72.1 & 95.1 & 96.6 & 97.1\\
					4 & 92.4 & 87.0 & 86.6 & 86.5 & 92.4 & 96.8 & \textbf{97.6} & \textbf{97.9}\\
					6 & 92.8 & 87.2 & 86.8 & 86.7 & 92.8 & 96.7 & 97.4 & 97.8\\
					8 & \textbf{92.9} & \textbf{87.4} & \textbf{87.0} & \textbf{86.9} & \textbf{92.9} & \textbf{96.9} & 97.5 & 97.8\\
					\hline
				\end{tabular}
			\end{small}
		\end{center}
		\vskip -0.1in
		
		\label{tab:length}
		\vskip -0.2in
	\end{table}

\noindent \textbf{The effect of autoregressive decoder.}
One natural baseline is to directly apply beam search to the prefix tree derived from RQ codes learned by the tokenizer, rather than remodeling the semantic relationship between RQ codes through the autoregressive decoder.  The comparison results on ImageNet dataset are summarized in \cref{tab:prefixtree}. The performance of this baseline is significantly inferior to our proposed method, highlighting the importance of our autoregressive decoder.

\begin{table}[h]
	\centering
    \centering
    \setlength\tabcolsep{5pt}
		\footnotesize
  \caption{Ablation study on the autoregressive decoder.}
\label{tab:prefixtree}
 \vskip -0.1in
		\begin{tabular}{l|c}
\toprule
Method & MAP@100\\
\hline
RQ prefix-tree & 56.7\\
IRGen(Ours) & \textbf{76.0}\\
 \hline
\end{tabular}
\vskip -0.2in

\end{table}

\noindent \textbf{The effect of reconstruction loss.}
During the training of image identifier, we propose to utilize M levels of partial reconstruction loss besides the traditional classification loss, as shown in Equation (3). \cref{tab:reconstructionloss} ablates this reconstruction loss and shows that this loss is beneficial for learning semantic image identifier.

\begin{table}[h]
\begin{center}
\setlength\tabcolsep{5pt}
\begin{footnotesize}
\caption{Ablation study on the reconstruction loss during the training of image identifiers.}
\label{tab:reconstructionloss}
\vskip -0.2in
\begin{tabular}{l|cccc}
			\toprule
			\multirow{2}{*}{Loss} & \multicolumn{4}{c}{Precision} \\
			\cline{2-5}
			& 1 & 10 & 20 & 30\\
			\hline
			Only classification loss & 92.1 & 85.1  & 83.4 & 82.5\\
			Full loss & \textbf{92.4} & \textbf{87.0} & \textbf{86.6} & \textbf{86.5}\\
			\hline
		\end{tabular}
\end{footnotesize}
\end{center}
\vspace{-1em}

\vskip -0.2in
\end{table}

	\section{Qualitative Retrieval Results}
\begin{figure*}
		\centering
		\includegraphics[width=.85\linewidth, clip]{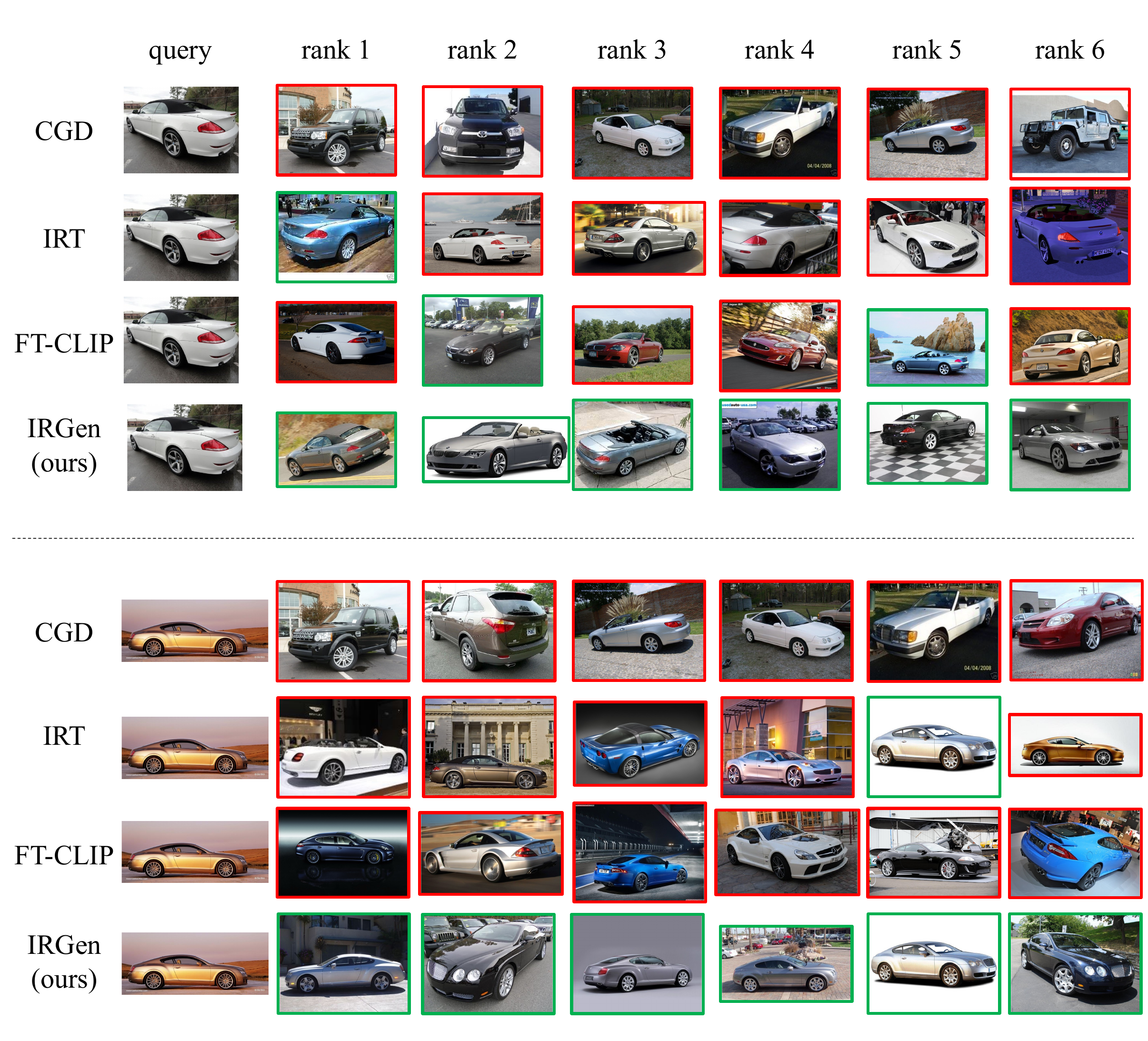}
		\caption{Examples on Cars196 dataset.Results of CGD, IRT, FT-CLIP, our IRGen are shown from top to bottom.The results of CGD, IRT, FT-CLIP are retrieved by SPANN.}
		\label{fig:cars}
	\end{figure*}

\begin{figure*}
		\centering
		\includegraphics[width=.85\linewidth, clip]{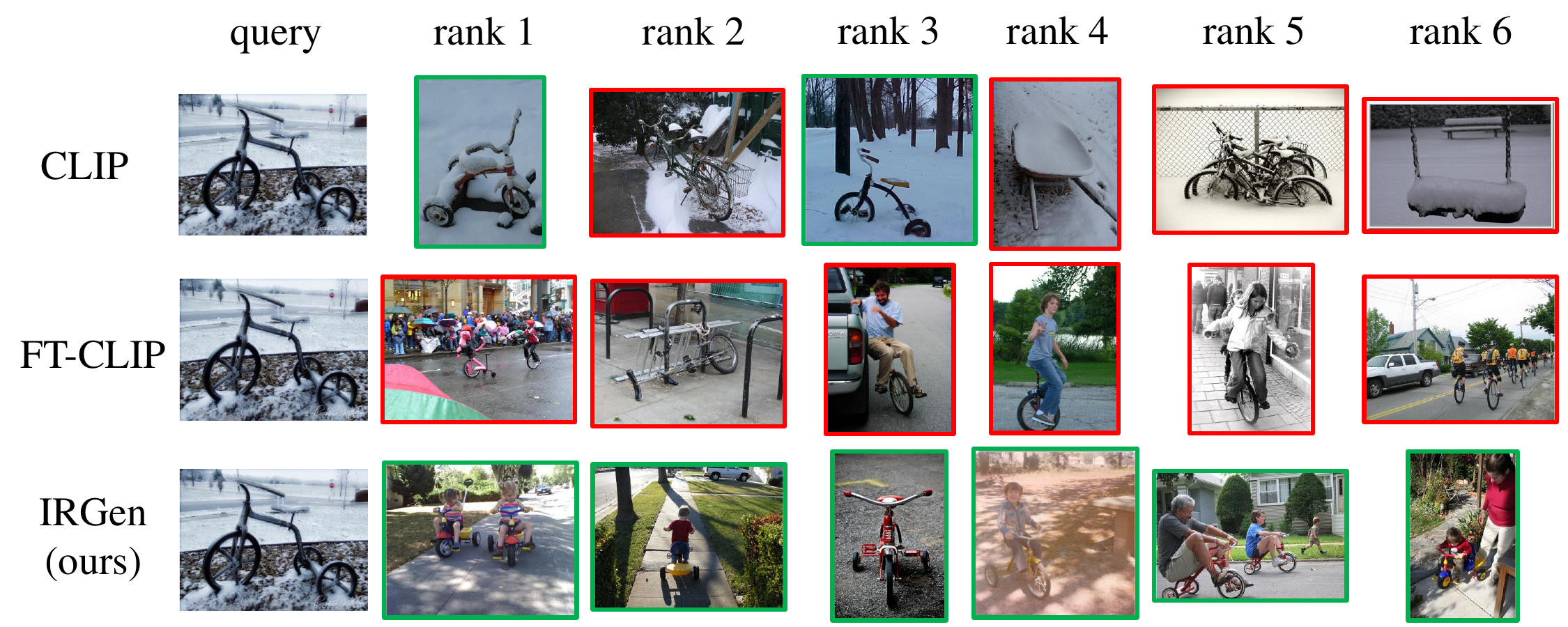}
		\caption{Examples on ImageNet dataset.Results of CLIP, FT-CLIP, our IRGen are shown from top to bottom. The results of CLIP, FT-CLIP are retrieved by SPANN.}
		\label{fig:imagenet}
	\end{figure*}

\begin{figure*} [t]
		\centering
		\includegraphics[width=0.8\linewidth, clip]{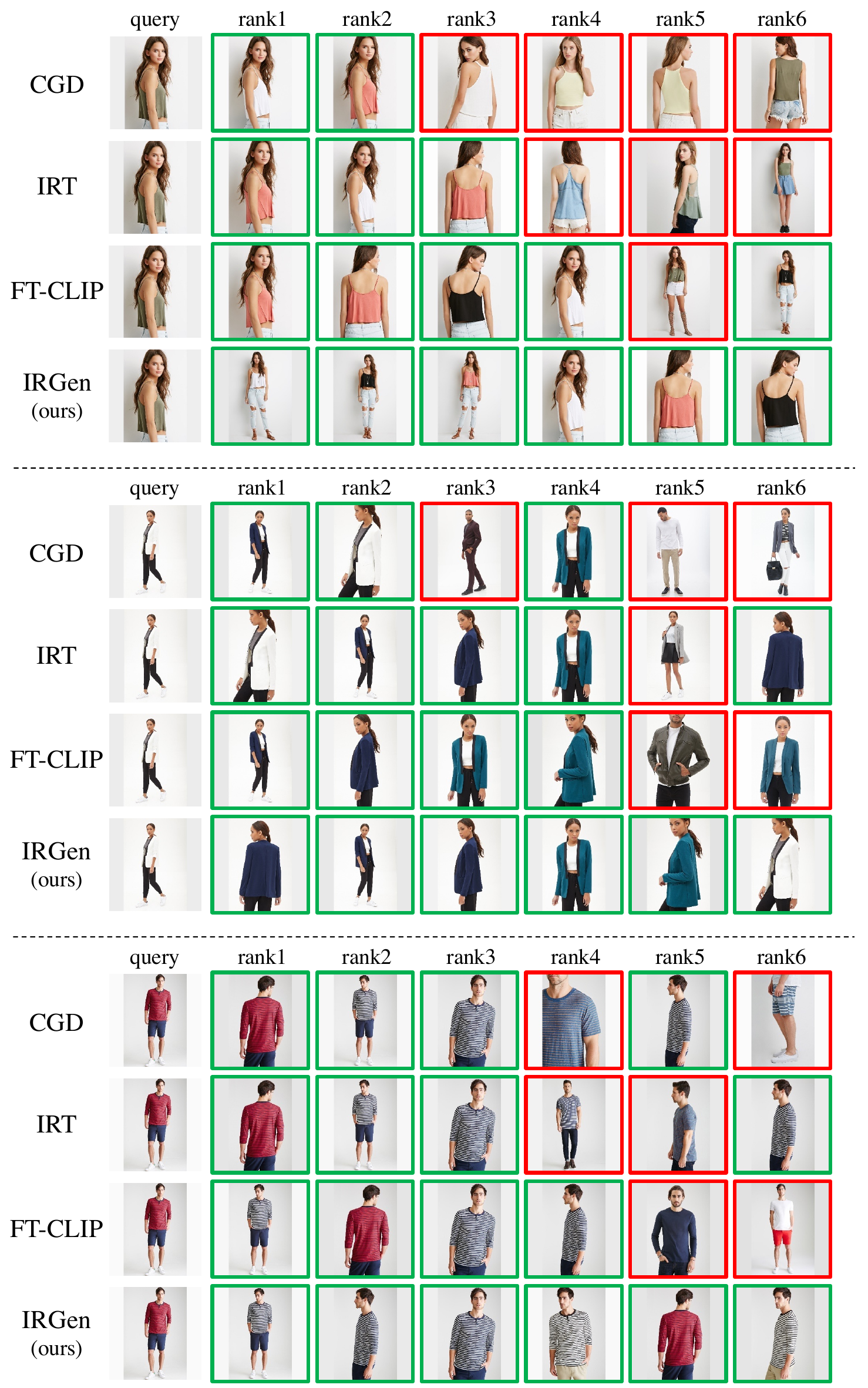}
		\caption{Examples on In-shop Clothes dataset.Results of CGD, IRT, FT-CLIP, our IRGen are shown from top to bottom.The results of CGD, IRT, FT-CLIP are retrieved by SPANN.}
		\label{fig:inshop}
	\end{figure*}

	In this section, we provide several retrieval examples that showcase the performance of our approach compared to baselines. The retrieval results on In-shop Clothes, Cars196, and ImageNet using different methods are depicted in \cref{fig:inshop}, \cref{fig:cars}, and \cref{fig:imagenet}, respectively. Correctly retrieved images are highlighted with green borders, while incorrectly retrieved ones are marked with red borders.
	Upon examining the results presented in these figures, it becomes evident that our proposed method performs exceptionally well and is capable of handling even the most challenging examples.
	
\end{document}